\documentclass[letterpaper]{article} 
\usepackage{aaai24}  
\usepackage{times}  
\usepackage{helvet}  
\usepackage{courier}  
\usepackage[hyphens]{url}  
\usepackage{graphicx} 
\usepackage{natbib}  
\usepackage{caption} 
\usepackage{algorithm}
\usepackage{algorithmic}

\usepackage[table,xcdraw]{xcolor}
\usepackage{booktabs}
\usepackage{bbding}
\usepackage{amssymb}
\usepackage{amsmath}
\usepackage{multirow}
\usepackage{soul}

\usepackage{newfloat}
\usepackage{listings}
\DeclareCaptionStyle{ruled}{labelfont=normalfont,labelsep=colon,strut=off} 
\floatstyle{ruled}
\newfloat{listing}{tb}{lst}{}
\floatname{listing}{Listing}
\title{KeDuSR: Real-World Dual-Lens Super-Resolution via Kernel-Free Matching}
\author{
    Huanjing Yue\textsuperscript{\rm 1},
    Zifan Cui\textsuperscript{\rm 1}, Kun Li\textsuperscript{\rm 2}, Jingyu Yang\textsuperscript{\rm 1}\thanks{Corresponding author.}
}
\affiliations{
    \textsuperscript{\rm 1}School of Electrical and Information Engineering, Tianjin University, China\\
    \textsuperscript{\rm 2}College of Intelligence and Computing, Tianjin University, China\\

    \{huanjing.yue, cuizifan, lik, yjy\}@tju.edu.cn

}

\usepackage{bibentry}

\begin{document}

\maketitle

\begin{abstract}
Dual-lens super-resolution (SR) is a practical scenario for reference (Ref) based SR by utilizing the telephoto image (Ref) to assist the super-resolution of the low-resolution wide-angle image (LR input). Different from general RefSR, the Ref in dual-lens SR only covers the overlapped field of view (FoV) area. However, current dual-lens SR methods rarely utilize these specific characteristics and directly perform dense matching between the LR input and Ref. Due to the resolution gap between LR and Ref, the matching may miss the best-matched candidate and destroy the consistent structures in the overlapped FoV area. Different from them, we propose to first align the Ref with the center region (namely the overlapped FoV area) of the LR input by combining global warping and local warping 
to make the aligned Ref be sharp and consistent. Then, we formulate the aligned Ref and LR center as value-key pairs, and the corner region of the LR is formulated as queries. In this way, we propose a kernel-free matching strategy by matching between the LR-corner (query) and LR-center (key) regions, and the corresponding aligned Ref (value) can be warped to the corner region of the target. Our kernel-free matching strategy avoids the resolution gap between LR and Ref, which makes our network have better generalization ability. In addition, we construct a DuSR-Real dataset with (LR, Ref, HR) triples, where the LR and HR are well aligned. Experiments on three datasets demonstrate that our method outperforms the second-best method by a large margin. \textit{Our code and dataset are available at https://github.com/ZifanCui/KeDuSR.}
\end{abstract}

\section{Introduction}

Single Image Super-Resolution (SISR)~\cite{liang2021swinir, yang2023two} aims to reconstruct a high-resolution (HR) image from a low-resolution (LR) input, which is challenging due to the limited available information. In contrast, Reference-based SR (RefSR) introduces a similar high-resolution reference image (Ref) to assist the reconstruction process and has achieved better results. However, the development of RefSR is constrained since obtaining similar Ref for a given LR in real scenarios is difficult. 

Fortunately, modern smartphones are equipped with multiple cameras of different fields of view (FoV), where the wide-angle lens sacrifices resolution to increase the FoV, while the telephoto lens has a smaller FoV but higher resolution. Therefore, dual-lens (or dual-camera) SR is proposed \cite{wang2021dual}, where the telephoto camera serves as the reference to super-resolve the wide-angle camera by transferring the matched reference details to the LR, as shown in Fig. \ref{fig:dataset}. However, only the center region (namely the overlapped FoV area) of the LR image has reference content, with differences in viewpoints and colors. Meanwhile, it is difficult for the corner region \footnote{We divide the LR image into two parts, where the overlapped FoV region is named as the center region and the remaining regions in the LR are named as corner region.} of the LR image to find similar contents from the reference due to the large resolution gap (as shown in Fig. \ref{fig:dataset}) between the telephoto and wide-angle cameras. Therefore, the key question for real dual-lens SR is how to improve the matching and warping performance between LR and Ref when they have large resolution gaps and different FoV?  

The matching problem has been widely explored in RefSR and dual-lens SR. Previous RefSR methods \cite{yang2020learning,lu2021masa,jiang2021robust} are conducted with synthesized LR (namely the down-sampled version of the HR) images from CUFED5 dataset, and the matching is conducted between HR$\downarrow$ (namely LR) and Ref$\downarrow$ (or Ref). However, when the HR and Ref are captured with different focal lengths, the resolution gap still exists. Similarly, DCSR \cite{wang2021dual} also utilizes synthesized pairs, namely that the original LR and the reference image are downsampled to generate the training triples, namely \{LR$\downarrow$, Ref$\downarrow$, LR\}, where the original LR serves as the ground truth. The matching is conducted between LR$\downarrow$ and Ref$\downarrow\downarrow$. However, the simple downsampling operation cannot simulate the resolution gap between LR and Ref since they are captured by different focal lengths, as shown in Fig. \ref{fig:dataset}. SelfDZSR \cite{zhang2022self} directly performs matching between Ref and auxiliary-LR features, which also did not pay attention to the resolution gap. Different from them, we did not perform matching between the features of Ref$\downarrow$ and LR. We propose a kernel-free matching strategy by matching between LR-corner and LR-center features, which perfectly avoids the resolution gap problem. In this way, our method also has a good generalization ability.  

The warping strategy is the second key problem for dual-lens SR. General RefSR, such as \cite{yang2020learning, huang2022task} performs globally pixel-wise or patch-wise dense matching since the corresponding matched content may locate in any position in the reference and then utilize the matched index for Ref warping. The benchmark dual-lens SR work DCSR \cite{wang2021dual} also utilizes this strategy. However, for dual-lens SR, the center region of the LR has the same scene as that of the reference. Directly performing dense patch matching between the LR and Ref may miss the best-matched patch for the center region due to the large resolution gap between the LR and Ref and the matching index may be incongruent. Therefore, SelfDZSR \cite{zhang2022self} proposes to paste the Ref back to the center area of the warped Ref features. However, the LR center and Ref are not pixel-wise aligned and this operation introduces misalignments between the center and corner regions in the SR result. Different from them, we propose a novel center warping strategy to find the matched content for the center region, which jointly utilizes global warping and local warping. For the corner region, we utilize the kernel-free matching index for corner warping, leading to a well-aligned reference in both corner and center regions.  

The third problem is how to adapt to real captured LR images in dual-lens SR? Since there is no pairwise real dual-lens SR dataset, DCSR \cite{wang2021dual} is trained with a synthesized dataset. It adapts the network to real images by finetuning with self-supervised loss. However, the finetuning strategy cannot solve the domain gap problem between synthesized and real captured ones. In contrast, SelfDZSR \cite{zhang2022self} proposes a self-supervised learning strategy, which utilizes the warped telephoto image as the ground truth (GT) and the center region of it serves as the reference. It introduces an auxiliary LR to make the warped LR and Ref aligned with GT during training. This makes the network pay more attention to the alignment process other than detail generation.
 
Recently, ZeDuSR \cite{xu2023zero} proposes zero-shot learning to deal with real captured LR images by training with center region pairs, but it requires a long inference time due to online learning. Different from them, we argue that a well-aligned dual-lens SR dataset is required to further boost the real dual-lens SR performance. Therefore, we construct the first well-aligned DuSR-Real dataset, where the HR is aligned with the real captured LR and the reference is also real captured. The LR and Ref have overlapped FoV regions. In addition, we reorganize the previous dual-lens SR datasets and construct another two real datasets, namely RealMCVSR-Real and CamereFusion-Real, for comprehensively evaluation.

In summary, our contributions are as follows.
\begin{itemize}
    \item We are the first to explore the real dual-lens SR problem via supervised learning. We propose a  center warping and corner warping strategy to align the reference with the LR input, which greatly improves the alignment quality. 
    \item We propose a kernel-free matching strategy by matching between LR-center and LR-corner, which avoids the resolution gap between the LR and Ref and makes the result be consistent across the whole image. 
    \item We constructed the first well-aligned DuSR-Real dataset. Extensive experiments on three datasets demonstrate the superiority of the proposed method. In addition, our method has the best generalization ability.   
\end{itemize}

\section{Related Work}
\label{sec:Related}


\subsection{Reference-Based SR}
RefSR, which leverages an HR reference to improve the SR performance, is a classical topic. From traditional methods to deep learning-based methods, the RefSR performance has been greatly improved. The key problem in RefSR is matching and warping. To improve the matching and warping performance, many sophisticated methods have been proposed, such as dense patch matching based \cite{yue2013landmark,zheng2017learning,zhang2019image, yang2020learning}, optical-flow based warping \cite{zheng2018crossnet}, and dense matching assisted DCN (deformable convolution network) warping, such as  \cite{shim2020robust, jiang2021robust, huang2022task}. Specifically, C2-matching~\cite{jiang2021robust} utilized contrastive learning to overcome scale and rotation transformation gaps and employed a teacher-student correlation distillation network to address the resolution gap. To accelerate the matching process, MASA~\cite{lu2021masa} and AMSA~\cite{xia2022coarse} explore efficient matching via a coarse-to-fine matching approach. Besides matching, advanced training and strategies are also emerging. RRSR~\cite{zhang2022rrsr} proposed a novel reciprocal training strategy. DATSR~\cite{cao2022reference} incorporated transformer into RefSR and have achieved SOTA performance. However, all these methods perform matching between LR and Ref. Different from them, we propose a kernel-free matching strategy tailored for dual-lens SR, by matching between LR-corner and LR-center regions, which avoids the resolution gap between the LR and Ref.

\subsection{Dual-Lens SR}
Compared with RefSR, dual-lens SR is more practical since the telephoto camera can directly serve as the reference for the wide-angle camera. DCSR ~\cite{wang2021dual} was the pioneer in introducing the dual-lens SR task. Since the training was conducted with a synthesized dataset, they further proposed a self-supervised domain adaptation strategy to generalize to real-world images. To enhance matching robustness, Zou \textit{et al.}~\shortcite{zou2023geometry} introduced geometric constraints to make the matching results be smooth. To adapt to real images, 
SelfDZSR~\cite{zhang2022self} proposed a self-supervised learning framework that directly utilized weakly aligned real-world pairs for training. ZeDuSR~\cite{xu2023zero} proposed a zero-shot learning strategy by training with the pairs inside the overlapped FoV region, which had a good generalization ability.  RefVSR~\cite{lee2022reference} and ERVSR \cite{kim2023efficient} extended the dual-lens SR strategy to video SR. 
Unlike them, we jointly utilize center and corner warping to improve the alignment performance between LR and Ref.

\subsection{Real-World SR Datasets}
The quality of the dataset is an important factor in promoting network development and improving SR performance. The widely used SR datasets are usually constructed by downsampling the GT, thus resulting in well-aligned LR-HR pairs, such as the SISR datasets, i.e., DIV2K \cite{timofte2017ntire} and the RefSR dataset CUFED5 \cite{zhang2019image}. However, the models trained on the synthesized dataset cannot generalize well to real degraded images. Therefore, many real-world SR datasets are collected by capturing with different focuses, such as City100~\cite{chen2019camera}, SR-Raw~\cite{zhang2019zoom}, DRealSR~\cite{wei2020component}, which have greatly improved the model's ability in dealing with real captured LR images. However, these datasets cannot be directly utilized for dual-lens SR due to the lack of triples (LR, Ref, and HR). The benchmark dual-lens SR datasets, i.e., CameraFusion~\cite{wang2021dual} and RealMCVSR~\cite{lee2022reference}, construct the training triples by downsampling the LR and Ref, generating \{LR$\downarrow$, Ref$\downarrow$, LR\}, where the original LR image serves as the GT. To deal with real LR images, SelfDZSR~\cite{zhang2022self} proposes to use the misaligned triples for training, which requires tedious operations to deal with the misalignment problem. Afterward, ZeDuSR \cite{xu2023zero} explores zero-shot learning to solve the real dual-lens SR. Different from them, we argue that a real triple dataset is still needed to further boost the development of real-world dual-lens SR. Therefore, we construct a DuSR-Real dataset with well-aligned LR and HR pairs and corresponding assisted HR references with overlapped FoV.

\begin{figure}[t]
  \centering

  \includegraphics[width=1\linewidth]{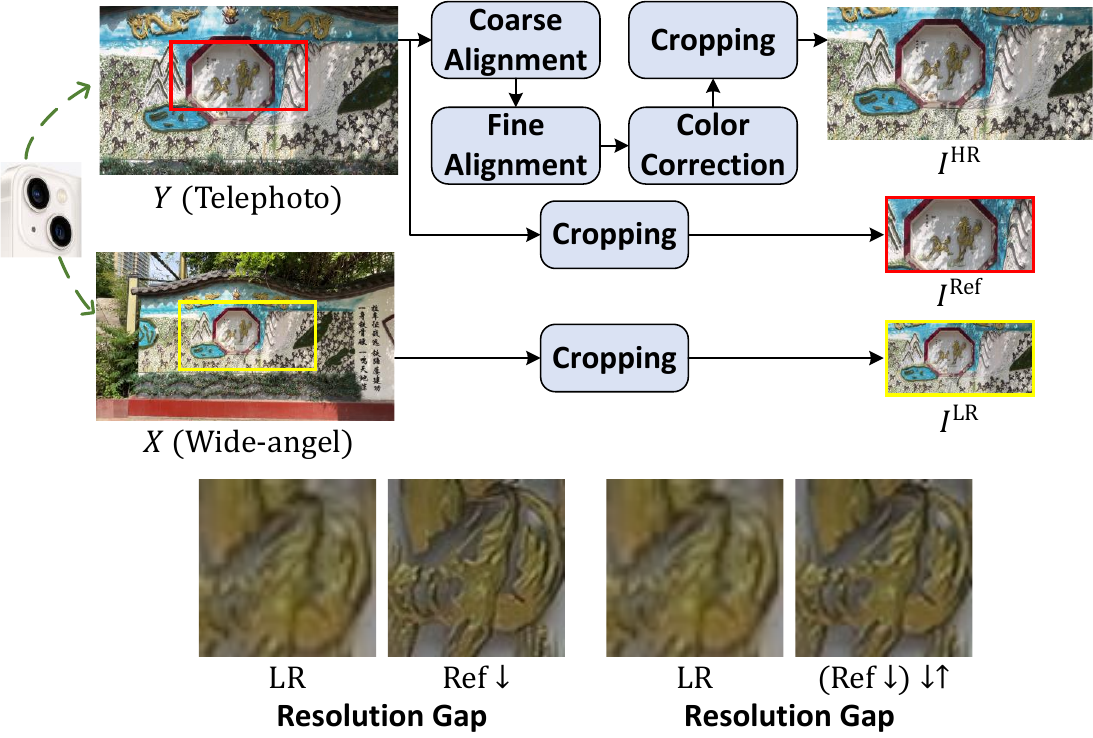}
   \caption{Illustration of our DuSR-Real dataset construction process and the resolution gap between LR and Ref. }
   \label{fig:dataset}
\end{figure}

\section{DuSR-Real Dataset Construction}
\label{sec:Dataset}
In the literature, there are two datasets, i.e., CameraFusion \cite{wang2021dual} and RealMCVSR~\cite{lee2022reference} for dual-lens SR. However, they only provide LR and Ref pairs, without HR ground truth for the LR. In this work, we construct real triples for dual-lens SR.  

\textbf{Data Collection.} 
The scene numbers in CameraFusion and RealMCVSR are relatively small. Therefore, we collect a new large dataset for real dual-lens SR. Specifically, we use an iPhone 13 to capture dual-lens images through the DoubleTake App \footnote{https://apps.apple.com/us/app/doubletake-by-filmic-pro/id1478041592}. The focal length of the telephoto lens is two times that of the wide-angle lens. 
Note that, different from CameraFusion~\cite{wang2021dual}, which utilizes lens switching to capture the same scene, we simultaneously activate both lenses for capturing. In this way, our dataset can avoid the misalignments between the two cameras in dynamic scenes and is consistent with real applications.


\textbf{Data Processing.}
For supervised learning, we need to generate the HR GT for the input LR. We propose to warp the telephoto image with the wide-angle image to generate the HR GT and the original center area of the telephoto image can serve as the Ref.   
We adopt the coarse-to-fine alignment strategy proposed in ~\cite{yue2022real} to create well-aligned LR-HR pairs. As depicted in Fig. \ref{fig:dataset}, 
$X$ and $Y$ represent the original images captured by the wide-angle lens and telephoto lens, respectively. Firstly, we employ SIFT~\cite{lowe2004distinctive} and RANSAC~\cite{fischler1981random} to calculate the optimal homography matrix to coarsely align Y with X. Then, we adopt Deepflow~\cite{weinzaepfel2013deepflow} for fine alignment. In this work, we focus on the SR task and the color differences between LR and HR will affect the learning process. Therefore, we further utilize color correction, namely a linear scaling coefficient for each channel ~\cite{yue2022real}, to make them have similar colors. Then, we crop the overlapped area between X and warped Y, generating the LR-HR pair $I^{\text{LR}}$ and $I^{\text{HR}}$. Then, we further crop the central area (according to the relative position between the two lenses) of Y, generating $I^\text{Ref}$ to serve as the reference. 
We totally captured 730 pairs, and manually removed 255 triples with alignment errors. Among the remaining triples ($I^{\text{LR}}$,$I^\text{Ref}$,$I^{\text{HR}}$), 420 triples are used for training, and 55 triples are used for testing.

In order to perform cross-dataset evaluation, we further apply the same processing approach on the CameraFusion~\cite{wang2021dual} and RealMCVSR~\cite{lee2022reference} datasets, to generate well-aligned real LR-HR pairs and the pairs that cannot be well aligned are removed. The reorganized datasets are named CameraFusion-Real and RealMCVSR-Real, respectively.  Detailed information about the three datasets is provided in our supp. file.

\section{Method}

\subsection{Framework Overview}
Dual-lens SR is different from general RefSR since the Ref in dual-lens SR shares the same scene with that of the LR center. Therefore, we propose to deal with the center and corner region differently. As shown in Fig. \ref{fig:KeDuSR}, the Ref image first goes through center warping, and then with the index obtained from kernel-free matching, we further perform corner warping on the Ref. Combining the warped reference feature and the LR feature via an adaptive fusion module and reconstruction module, we obtain the final result $I^\text{SR}$. 
The following gives details about these modules in our Kernel-free matching based Dual-lens SR (termed as KeDuSR).

\begin{figure}[t]
  \centering
  \includegraphics[width=1\linewidth]{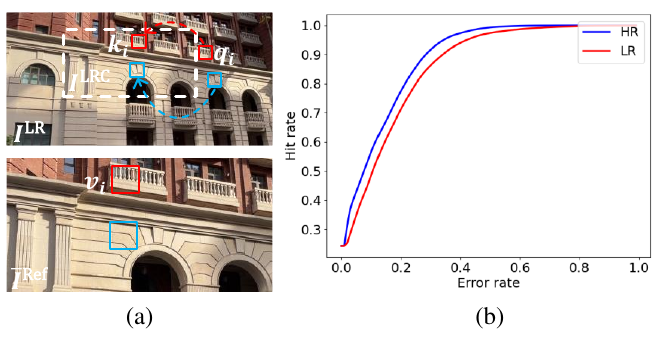}
   \caption{Illustration of the similarity between the corner and center regions, where (a) shows the similar patch pairs and (b) is the matching curve. The LR-center region is circled by a white dotted box and $\bar{I}^\text{Ref}$ is its corresponding HR Ref.}
   \label{fig:matching}
\end{figure}


\subsection{Center Warping}\label{sec:Center-warping}
Given the LR image $I^\text{LR}$ and its reference $I^\text{Ref}$, we need to first identify the overlapped area between them. Specifically, we utilize SIFT \cite{lowe2004distinctive} matching to find the matched points between $I^\text{Ref}$ and $I^\text{LR}$. Then, we utilize RANSAC \cite{fischler1981random} to filter outliers, and use the inliers to calculate the homography matrix, which is applied on $I^\text{Ref}$ and generates the warped reference $\bar{I}^\text{Ref}$. Afterward, we crop the area in $I^\text{LR}$ that corresponds to $\bar{I}^\text{Ref}$, and name it as the center region of the LR, denoted as $I^{\text{LRC}}$. After this global warping, $\bar{I}^\text{Ref}$ is coarsely aligned with $I^{\text{LRC}}$.

However, there are still small displacements between $\bar{I}^\text{Ref}$ and $I^{\text{LRC}}$, which may cause alignment errors in the following corner warping. Therefore, we further utilize the faster and differentiable flow-guided DCN \cite{chan2022basicvsr++} for local warping. Note that, we did not utilize DeepFlow for center warping since it is much slower and nondifferentiable. To reduce the computation cost, we downsample $\bar{I}^\text{Ref}$ to make it have the same scale as that of $I^{\text{LRC}}$. Then, we utilize the pretrained Spynet~\cite{ranjan2017optical} to compute the optical flow $f$ between $\bar{I}^\text{Ref}$$\downarrow$ and $I^{\text{LRC}}$. Then, we utilize residual blocks (ResBlocks) to extract features from $I^{\text{LRC}}$ and $\bar{I}^\text{Ref}$, generating $F^{\text{LRC}}$ and $\bar{F}^\text{Ref}$. Afterwards, the upsampled optical flow $f$$\uparrow$ is utilized to guide DCN ~\cite{dai2017deformable, zhu2019deformable} to align $\bar{F}^\text{Ref}$ with $F^{\text{LRC}}$, generating the fine-aligned features $\hat{F}^\text{Ref}$.  

Note that, compared with the widely used dense patch matching-based warping, our global warping and local warping combined strategy can preserve the reference image structures, which not only improves the following corner-warping performance but also improves the final SR quality in the center region. 


\subsection{Kernel-Free Matching and Corner Warping}\label{sec:Matching}
An intuitive strategy for warping the corner region is performing patch matching between the corner of $I^{\text{LR}}$ and the reference $I^{\text{Ref}}$. However, there is a large resolution gap between them. Even after downsampling, the resolution gap still exists between $I^{\text{LR}}$ and $I^{\text{Ref}}$$\downarrow\downarrow\uparrow$ (as shown in Fig. \ref{fig:dataset}) since simple downsampling cannot simulate the mapping between the two cameras. Another strategy is learning the mapping process via KernelGAN \cite{bell2019blind} or probabilistic degradation model (PDM) \cite{luo2022learning}, and utilizing the learned kernel to degrade $I^{\text{Ref}}$. However, the kernel depends on cameras, which makes the kernel learned with one specific camera does not generalize well to other cameras.

In contrast, we observe that due to the nonlocal similarity, for one query patch ($q_i$) in the corner region, we can find its similar patch ($k_i$) in the center region, as shown in Fig. \ref{fig:matching} (a). To visualize the similarity between corner region and center region, we plot the matching curve, namely hit rate versus error rate in Fig. \ref{fig:matching} (b). The error rate is defined as $e_r = \|q-k\|_2/\|q\|_2$, where $k$ is the matched patch from the center region with the minimal mean square error for $q$. The hit rate is the percentage of the query patch whose error rate is smaller than $e_r$. The blue matching curve is obtained by matching between the HR query and the HR key,  but the $e_r$ is obtained by the HR patches with the corresponding index. For more than 90\% of patches, their error rates are smaller than 0.3, which indicates a high similarity between corner and center regions. In addition, the matching in LR domain can well approximate the matching in HR domain. Therefore, in this work, we propose to perform matching between the corner and center regions of the LR image. Since this matching process is not influenced by different camera kernels, we formulate it as kernel-free matching.


Following \cite{yang2020learning,wang2021dual}, we also perform matching in the VGG (denoted as $\phi$) feature space by extracting features from $I^{\text{LR}}$ and $I^{\text{LRC}}$. The features are densely divided into $3\times3$ patches with a stride of 1, and the cosine similarity $S_{i,j}$ is computed between each patch pair, namely $P^{\text{LR}}_i$ and $P^{\text{LRC}}_j$. For  $P^{\text{LR}}_i$, its matched patch is the one that has the highest similarity score, and the matched index $M_i$ and confidence score $C_i$ can be obtained by 
\begin{equation}\label{index}
{M_i} = \mathop {\arg \max }\limits_j {S_{i,j}},{C_i} = \mathop {\max }\limits_j {S_{i,j}}. 
\end{equation}
Then, we utilize the matched index map to extract HR matched patches from the Ref. Therefore, the reference warping result for the corner region is 
\begin{equation}
    \tilde{F}^{\text{Ref}}_{2i} = \hat{F}^{\text{Ref}}_{M_i},
\end{equation}
where $\tilde{F}^{\text{Ref}}_i$ denotes the patch value of $\tilde{F}^{\text{Ref}}$ in the $i^\text{th}$ patch position. Since the query patches are densely extracted, the overlapped value patches are averaged in the overlapped region. In addition, the center region of $\tilde{F}^{\text{Ref}}$ is indeed the center warping result $\hat{F}^{\text{Ref}}$. Correspondingly, the center region of the confidence map is set to 1.

\begin{figure*}[t]
  \centering

  \includegraphics[width=1.0\linewidth]{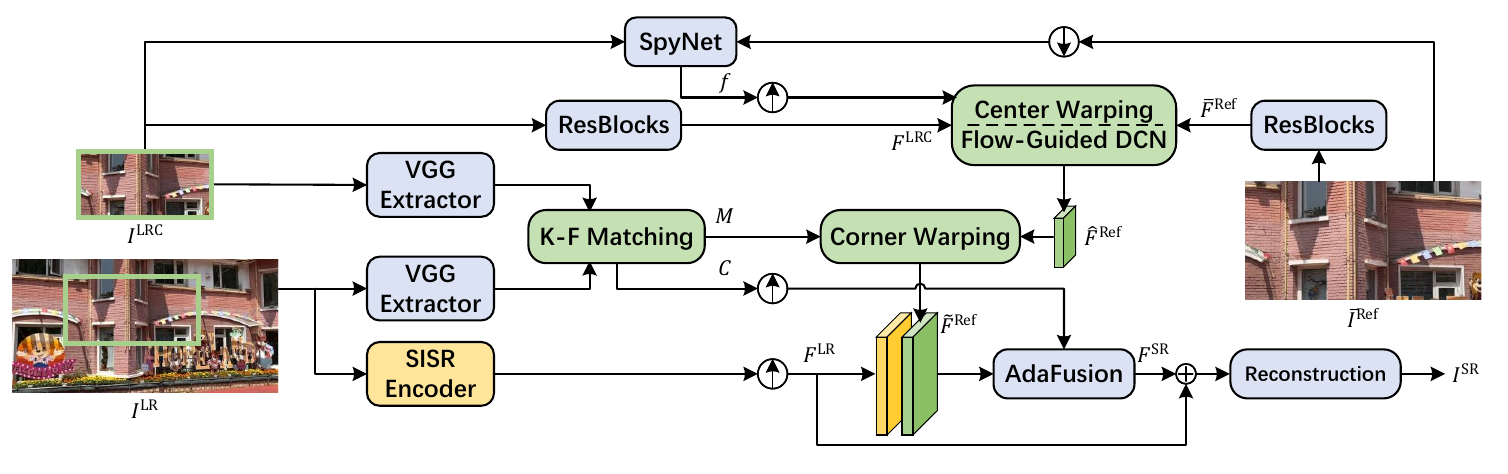}
   \caption{The framework of our KeDuSR. Kernel-Free Matching is performed between LR-corner and LR-center(\(I^\text{LRC}\)) to obtain the index map \(M\) and confidence map \(C\). Then, employing center warping and corner warping,  we obtain the warped high-resolution feature map \(\tilde{F}^\text{Ref}\) of the reference. After fusion with \(F^\text{LR}\), we generate the SR result \(I^\text{SR}\).}
   \label{fig:KeDuSR}
\end{figure*}

\subsection{SISR Encoder}
As demonstrated in \cite{huang2022task}, coupling the image super-resolution task from the input LR image with the texture transfer task from the reference will introduce interference. Therefore we also decouple the SISR task into a separate module, which is constructed by 24 residual blocks with channel attention. The extracted features are upsampled (named as $F^\text{LR}$) to cope with the size of the aligned reference feature. Note that, different from the two-stage training approach used in ~\cite{huang2022task}, we train our SISR encoder as an integral part of the entire network. 




\subsection{Adaptive Fusion}
The warped reference feature $\tilde{F}^{\text{Ref}}$ and LR image feature $F^\text{LR}$ are complementary to each other, and we utilize a fusion model to fuse them together. Since the matching quality for different positions is different, inspired by \cite{wang2021dual},  we also utilize adaptive fusion by introducing the confidence map obtained in the matching process. In addition, instead of fusing $\tilde{F}^{\text{Ref}}$, we fuse its high-frequency part $\tilde{F}^{\text{Ref}}_{hf} = \tilde{F}^{\text{Ref}} - \tilde{F}^{\text{Ref}}\downarrow\uparrow$. This process is formulated as  
\begin{equation}\label{fusion}
{F^{SR}} = \Phi(\text{concat}(g(C) \cdot \tilde{F}^\text{Ref}_{hf},F^{LR})),
\end{equation}
where \(C\) denotes the confidence map, $g()$ represents the  convolution operations, $\cdot$ represents the element-wise multiplication, and $\Phi$ represent the AdaFusion block. The AdaFusion module is constructed by ResBlocks with spatial and channel attention.

\subsection{Loss Functions}
Following previous RefSR methods, we also utilize hybrid loss functions. First, for reconstruction loss, we utilize Charbonnier loss~\cite{lai2017deep}, which is a differentiable variant of $\ell_1$ loss, denoted as 
\begin{equation}\label{loss}
\mathcal{L}_{ch} = \sqrt {\parallel I^\text{HR} - I^\text{SR}\parallel _2^2 + \varepsilon },
\end{equation}
where \(\varepsilon  = 1 \times {10^{ - 6}}\). \(I^\text{HR}\) denotes the HR GT, while \(I^\text{SR}\) denotes the SR result.
For better visual effects, we further incorporate perceptual loss and adversarial loss. 
The perceptual loss is expressed as 
\begin{equation}
{\mathcal{L}_{per}} = \parallel {\phi _i}({I^\text{HR}}) - {\phi _i}({I^\text{SR}}){\parallel _2},
\end{equation}
where \(\phi _i\) denotes the \(i\)-th layer of VGG19. 
We adopt the Relativistic GANs~\cite{jolicoeur2018relativistic} as our adversarial loss~\cite{goodfellow2014generative}, denoted as \(\mathcal{L}_{adv}\). In summary, our hybrid loss can be represented as
\begin{equation}
\mathcal{L} = {\mathcal{L}_{ch}} + {\lambda _1}{\mathcal{L}_{per}} + {\lambda _2}{\mathcal{L}_{adv}}, 
\end{equation}
where the weighting parameters \(\lambda _1\) and \(\lambda _2\) are set to \(1 \times{10^{ - 3}}\) and \(1 \times{10^{ - 4}}\), respectively. Note that, we provide two results in experiments. One is trained with only the reconstruction loss $\mathcal{L}_{ch}$ and the other is trained with the hybrid loss $\mathcal{L}$. 



\section{Experiments}
\label{sec:Experiments}

\subsection{Training Details and Datasets}
During training, the batch size is 4, and the patch size for the input LR is $128\times128$. We utilized the Adam optimizer~\cite{kingma2014adam} and the cosine annealing scheme~\cite{loshchilov2016sgdr}. The learning rate is initially set to \(10^{ - 4}\) and is decayed to \(10^{ - 6}\). All experiments were conducted using PyTorch~\cite{paszke2019pytorch} on an Nvidia GeForce RTX 3090 GPU.

We conduct comparisons on three datasets, namely our DuSR-Real, the reorganized CameraFusion-Real, and RealMCVSR-Real datasets. Our DuSR-Real contains 420 training triples and 55 testing images, and the HR GT has a resolution of $1792\times896$. The RealMCVSR-Real (CameraFusion-Real) consists of 330 (83) training triples and 50 (15) testing images, and the GT has a resolution of $1792\times896$ ($3584\times2560$). 

\subsection{Comparison with State-of-the-arts}
To evaluate the effectiveness of our KeDuSR, we compare with three kinds of SR methods, including the SISR methods: RCAN~\cite{zhang2018image}, SwinIR~\cite{liang2021swinir}, ESRGAN~\cite{wang2018esrgan}, BSRGAN~\cite{zhang2021designing}, the RefSR methods: TTSR~\cite{yang2020learning}, MASA~\cite{lu2021masa}, DASTR~\cite{cao2022reference}, and the dual-lens SR methods: DCSR~\cite{wang2021dual}, SelfDZSR~\cite{zhang2022self}, ZeDuSR \cite{xu2023zero}. For a fair comparison, we retrained the aforementioned methods with the same training set as that used in our method.
SISR methods use the LR-HR pair during training, while RefSR methods use the LR-Ref-HR triples (except for ZeDuSR, which uses LR-Ref pairs). For SelfDZSR, since the Ref and LR have large and irregular displacements in our dataset, we did not paste the center Ref back to its warped features to avoid misalignment artifacts. 

\begin{table}[t]
  \centering

  \resizebox{\linewidth}{!}{

\begin{tabular}{lcccc}
\hline
\multirow{2}{*}{Method} & \multirow{2}{*}{\begin{tabular}[c]{@{}c@{}}Latency\\ (s)\end{tabular}} & Full-Image         & Center-Image & Corner-Image \\
                        &                                                                        & PSNR\(\uparrow\)/SSIM\(\uparrow\)/LPIPS\(\downarrow\)    & PSNR/SSIM    & PSNR/SSIM    \\ \hline\hline

RCAN-\(\ell\)                    & 0.69                                                                   & 26.44 / 0.8676 / 0.147 & 26.91 / 0.8704 & \underline{26.33} / \underline{0.8667} \\

SwinIR-\(\ell\)                  & 2.85                                                                    & 26.14 / 0.8601 / 0.157 & 26.35 / 0.8612 & 26.11 / 0.8597 \\

ESRGAN                  & 0.08                                                                   & 25.78 / 0.8622 / 0.152 & 25.91 / 0.8637 & 25.77 / 0.8617 \\

BSRGAN                  & 0.45                                                                   & 24.77 / 0.8227 / 0.202 & 25.01 / 0.8233 & 24.71 / 0.8225 \\ \hline

TTSR-\(\ell\)                    & 7.51                                                                    & 26.48 / 0.8676 / 0.147 & 27.69 / 0.8810 & 26.17 / 0.8631 \\

MASA-\(\ell\)                    & 1.52                                                                    & 26.36 / 0.8592 / 0.160 & 26.85 / 0.8620 & 26.25 / 0.8582 \\

DATSR-\(\ell\)                   & 9.35                                                                   & 26.17 / 0.8583 / 0.157 & 26.48 / 0.8596 & 26.11 / 0.8579 \\ \hline

DCSR-\(\ell\)                    & 0.84                                                                    & \underline{26.77} / \underline{0.8748} / 0.134   & \underline{28.87} / \underline{0.9078} & 26.29 / 0.8635 \\
DCSR                & 0.84                                                                    & 26.19 / 0.8553 / 0.110   & 28.05 / 0.8929 & 25.75 / 0.8425 \\

SelfDZSR-\(\ell\)                & 0.17                                                                    & 26.27 / 0.8559 / 0.158 & 26.97 / 0.8591 & 26.10 / 0.8548 \\
SelfDZSR            & 0.17                                                                    & 25.98 / 0.8455 / \underline{0.105} & 26.61 / 0.8496 & 25.81 / 0.8442 \\
ZeDuSR-\(\ell\)            & 180.0                                                                    & 25.41 / 0.8247 / 0.191 & 26.29 / 0.8336 & 25.21 / 0.8216 \\\hline\hline

KeDuSR-\(\ell\)               & 0.51                                                                    & \textbf{27.66} / \textbf{0.8890} / 0.117 & \textbf{29.58} / \textbf{0.9303} & \textbf{27.24} / \textbf{0.8750} \\
KeDuSR           & 0.51                                                                    & 27.18 / 0.8752 / \textbf{0.084} & 29.06 / 0.9219 & 26.77 / 0.8593 \\ \hline
\end{tabular}
  }
  \caption{ Quantitative comparisons on DuSR-Real. Bold and underlined indicate the best and second-best performance, respectively. -\(\ell\) denotes training with only reconstruction loss. Latency indicates the time required to generate one HR result ($1792\times896$) using one NVIDIA 3090 GPU.}
  \label{tab:DuSR-Real}
\end{table}

\begin{table}
  \centering

  \resizebox{\linewidth}{!}{
\begin{tabular}{lccc}
\hline
\multirow{2}{*}{Method} & Full-Image         & Center-Image & Corner-Image \\
                        & PSNR\(\uparrow\)/SSIM\(\uparrow\)/LPIPS\(\downarrow\)    & PSNR/SSIM    & PSNR/SSIM    \\ \hline\hline
 
RCAN-\(\ell\)                    & 25.96 / 0.8033 / 0.234 & 25.69 / 0.7937 & \underline{26.12} / \underline{0.8065} \\
 
SwinIR-\(\ell\)                  & 25.78 / 0.7982 / 0.246 & 25.50 / 0.7885 & 25.94 / 0.8015 \\ \hline
 
TTSR-\(\ell\)                    & 25.92 / 0.8017 / 0.235 & 25.94 / 0.7962 & 25.98 / 0.8036 \\
 
MASA-\(\ell\)                    & 25.95 / 0.7989 / 0.239 & 25.81 / 0.7899 & 26.07 / 0.8020 \\
 
DATSR-\(\ell\)                   & 25.81 / 0.7975 / 0.242 & 25.58 / 0.7882 & 25.95 / 0.8007 \\ \hline
 
DCSR-\(\ell\)                    & \underline{26.28} / \underline{0.8111} / 0.217 & \underline{27.19} / 0.8298 & 26.08 / 0.8048 \\
DCSR                & 25.85 / 0.7966 / 0.186 & 26.98 / \underline{0.8476} & 25.58 / 0.7793 \\
 
SelfDZSR-\(\ell\)                & 25.33 / 0.7928 / 0.246 & 25.66 / 0.7860 & 25.30 / 0.7952 \\
SelfDZSR            & 25.24 / 0.7786 / \underline{0.175} & 25.50 / 0.7732 & 25.23 / 0.7805 \\ 
ZeDuSR-\(\ell\)        & 24.98 / 0.7702 / 0.262 & 25.38 / 0.7650 & 24.93 / 0.7720 \\ \hline\hline
 
KeDuSR-\(\ell\)               & \textbf{27.05} / \textbf{0.8406} / 0.180 & \textbf{29.25} / \textbf{0.9191} & \textbf{26.56} / \textbf{0.8139} \\
KeDuSR           & 26.42 / 0.8184 / \textbf{0.127} & 28.51 / 0.9090 & 25.95 / 0.7875 \\ \hline
\end{tabular}
  }
\caption{Quantitative comparisons on RealMCVSR-Real.}
  \label{tab:RealMCVSR-Real}
\end{table}

\begin{table}
  \centering

  \resizebox{\linewidth}{!}{
\begin{tabular}{lccc}
\hline
                         & Full-Image         & Center-Image & Corner-Image \\
\multirow{-2}{*}{Method} & PSNR\(\uparrow\)/SSIM\(\uparrow\)/LPIPS\(\downarrow\)    & PSNR/SSIM    & PSNR/SSIM    \\ \hline\hline
 
RCAN-\(\ell\)                     & 25.67 / 0.8049 / 0.308 & 26.65 / 0.8158 & 25.45 / 0.8012 \\

SwinIR-\(\ell\)                   & 25.32 / 0.8007 / 0.315 & 25.81 / 0.8073 & 25.22 / 0.7985 \\ \hline

TTSR-\(\ell\)                     & 25.83 / 0.8044 / 0.311 & 26.75 / 0.8188 & 25.62 / 0.7996 \\
 
MASA-\(\ell\)                     & 25.78 / 0.8030 / 0.303 & 26.70 / 0.8155 & 25.58 / 0.7988 \\ \hline

DCSR-\(\ell\)                     & 26.02 / \underline{0.8123} / 0.293   & \underline{28.37} / \underline{0.8440} & 25.51 / \underline{0.8016} \\
DCSR                 & 25.47 / 0.7605 / 0.165   & 27.14 / 0.7883 & 25.08 / 0.7512 \\
DCSR-SRA      & 24.75 / 0.7347 / 0.189   & 25.62 / 0.7626 & 24.55 / 0.7254 \\
SelfDZSR-\(\ell\)                 & 25.94 / 0.8041 / 0.283 & 27.10 / 0.8148 & 25.68 / 0.8005 \\
SelfDZSR             & 25.64 / 0.7790 / \underline{0.151} & 26.77 / 0.7897 & 25.39 / 0.7753 \\ 
ZeDuSR-\(\ell\)        & \underline{26.16} / 0.7920 / 0.279   & 27.44 / 0.8067 & \underline{25.87} / 0.7871 \\ \hline\hline

KeDuSR-\(\ell\)                & \textbf{27.53} / \textbf{0.8292} / 0.276 & \textbf{30.48} / \textbf{0.8656} & \textbf{26.93} / \textbf{0.8169} \\
KeDuSR            & 27.00 / 0.7931 / \textbf{0.133} & 29.77 / 0.8418 & 26.43 / 0.7768 \\ \hline
\end{tabular}
  }
\caption{Quantitative comparisons on CameraFusion-Real.}
  \label{tab:CameraFusion-Real}
\end{table}

\textbf{Quantitative Comparison.}
We evaluate all the methods on three datasets, as shown in Tables \ref{tab:DuSR-Real}, \ref{tab:RealMCVSR-Real}, \ref{tab:CameraFusion-Real}. Full-image results represent the quantitative results of the entire image, the center-image corresponds to the results in overlapped FoV area, and the corner-image represents excluding the center-image from the full-image. For the models trained with only reconstruction loss, we denote it with $-\ell$. Otherwise, the corresponding model is trained with hybrid loss functions as proposed in their paper.   

On all three datasets, our method outperforms the second-best method by a large margin in terms of PSNR, SSIM \shortcite{wang2004image}, and LPIPS \shortcite{zhang2018unreasonable}. In addition, our method achieves the best performance in both the center and corner regions. 
For TTSR, its center result is better than that of RCAN due to the introduction of HR Ref. However, its corner result is worse since it cannot utilize Ref patches well. Meanwhile, DCSR works much better in the center region than TTSR due to its robust feature warping and fusion strategy. Different from them, we utilize a tailored center warping and kernel-free matching based corner warping, which greatly improves the matching performance in both center and corner regions. Meanwhile, the zero-shot learning (ZeDuSR) method can only utilize the single image information. Therefore, their performance is also inferior to ours. Note that, ZeDuSR works better in the CameraFusion-Real dataset since the image size in this dataset is large, which makes ZeDuSR extract more training pairs from one single image. In, addition, our method ranks second among all the RefSR methods in terms of latency.

We also evaluate the benefits of training with real pairs over finetuning. We utilize the released weights of DCSR, which is finetuned by the Self-supervised Real-image Adaptation (SRA) strategy, and term it DCSR-SRA. As shown in Table \ref{tab:CameraFusion-Real}, DCSR-SRA falls largely behind the original DCSR trained with our constructed real pairs.

\begin{figure*}[t]
  \centering

  \includegraphics[width=1\linewidth]{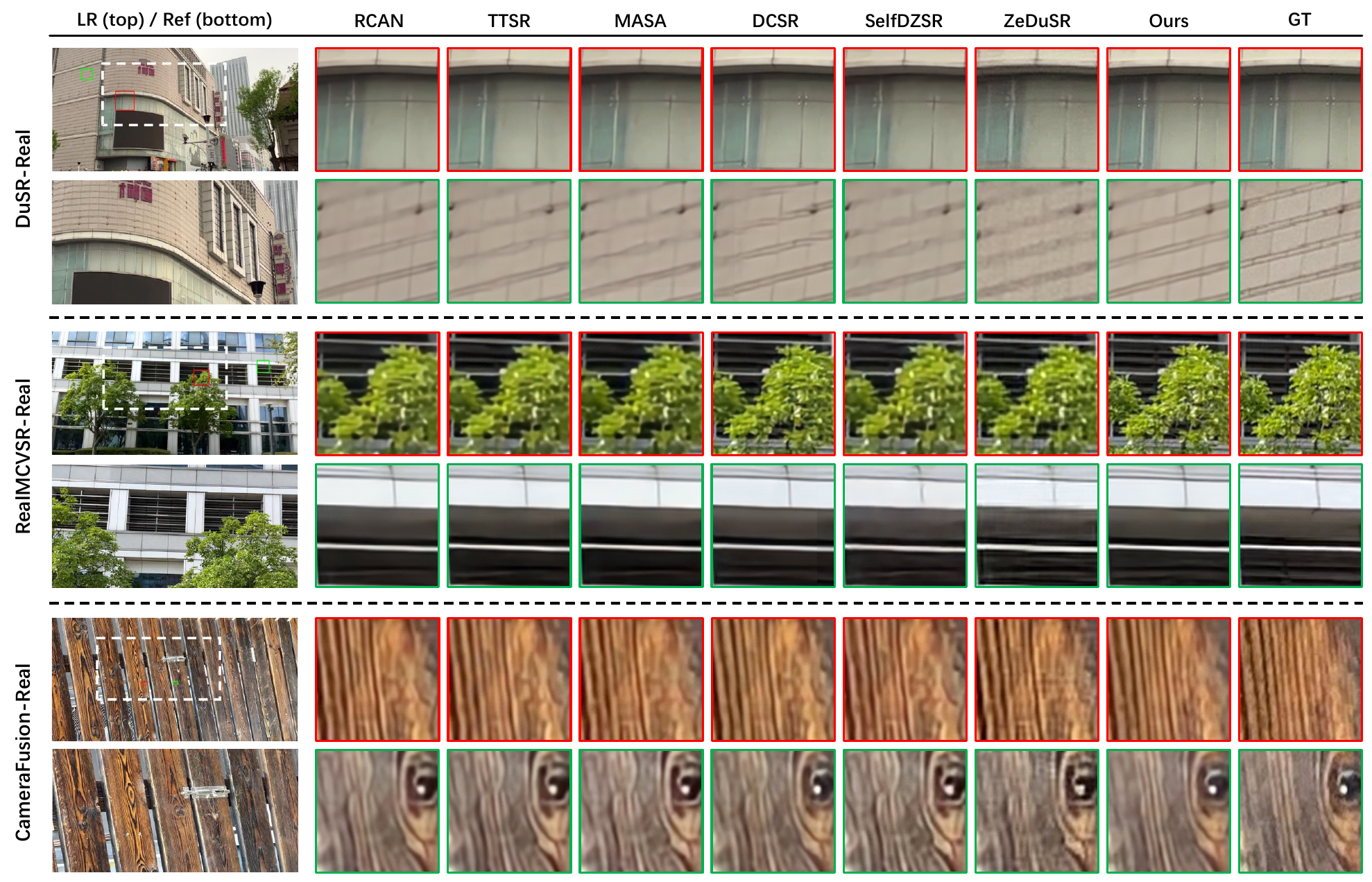}
   \caption{Visual comparisons on real-world dual-lens datasets. The white dotted box indicates the overlapped FoV area between LR and Ref. The presented results are obtained with only reconstruction loss.}
   \label{sec:visual}
\end{figure*}

\textbf{Qualitative Comparison.}
Fig. \ref{sec:visual} presents the visual comparison results. In the center region, our method achieves results that closely resemble the HR GT, surpassing other existing methods by a significant margin. In the corner region, our approach is capable of recovering more fine-grained details if similar textures are present in the Ref. In addition, for large LR input, the majority of methods need to divide the LR input into blocks due to the limitation of available memory. In this case, our method can avoid the blocking artifacts due to our kernel-free matching strategy while the compared methods suffer from these artifacts. More results are provided in our supp. file.

\begin{table}
  \centering
\resizebox{\linewidth}{!}{
\begin{tabular}{ccccc}
\hline
Variant & \begin{tabular}[c]{@{}c@{}}Traditional \\ Matching\end{tabular} & \begin{tabular}[c]{@{}c@{}}Center \\ Warping\end{tabular} & \begin{tabular}[c]{@{}c@{}}Corner\\ Warping\end{tabular} & PSNR\(\uparrow\)/SSIM\(\uparrow\)                           \\ \hline\hline
A       & \checkmark                                                               &                                                           &                                                          & 27.03 / 0.8801                        \\
B       &                                                                 & \checkmark                                                          &                                                         & 27.16 / 0.8757                        \\
C       &                                                                 &                                                        &  \checkmark                                                          & 27.50 / 0.8804                        \\ \hline \hline
D       &                                                                 & \checkmark                                                         & \checkmark                                                        & \textbf{ 27.66} / \textbf{0.8890} \\ \hline
\end{tabular}
}
\caption{Ablation study on our key modules, evaluated on DuSR-Real dataset.}
  \label{tab:Ablation}
\end{table}

\subsection{Ablation Study}
We conduct ablation experiments on our proposed matching and warping strategy. First, we remove the proposed center warping and corner warping and replace them with traditional dense feature matching, namely matching between the VGG features of $I^\text{LR}$ and $I^\text{Ref}\downarrow$. Then utilize the matching result for reference warping. As shown in Table \ref{tab:Ablation}, this (variant A) degrades the result by 0.63 dB. Note that, variant A is our baseline, which is better than SOTA RefSR methods, verifying that our proposed baseline is a robust baseline for the dual-lens SR task. For variant B, we remove the corner warping process, namely that we only utilize the center warping result and there is no reference for the corner region. Variant B still outperforms variant A in terms of PSNR since the center region is well reconstructed. This demonstrates that our global and local warping combined strategy is effective for the center region. For variant C, we remove the center warping, namely that we only utilize the coarsely aligned reference feature $\bar{F}^\text{Ref}$ for the center region, and thus the key-value pair in kernel-free matching is not accurately aligned. Therefore, the result of variant C is worse than our full model (variant D). In summary, our center warping and corner warping are essential for improving the dual-lens SR performance.

\subsection{Generalization Evaluation}
\begin{table}
  \centering

\resizebox{\linewidth}{!}{
\begin{tabular}{lcc}
\hline
\multirow{2}{*}{Method} & RealMCVSR-Real         & CameraFusion-Real      \\
                        & PSNR\(\uparrow\)/SSIM\(\uparrow\)/LPIPS\(\downarrow\)        & PSNR\(\uparrow\) / SSIM\(\uparrow\) / LPIPS\(\downarrow\)        \\ \hline
TTSR-\(\ell\)                    & 24.67 / 0.7814 / 0.248 & 25.23 / 0.7760 / 0.289 \\
MASA-\(\ell\)                    & 24.99 / 0.7830 / 0.258 & 25.45 / 0.7769 / 0.291 \\
DCSR-\(\ell\)                    & \underline{25.46} / \underline{0.7986} / \underline{0.226} & 25.58 / \underline{0.7931} / \underline{0.263} \\
SelfDZSR-\(\ell\)                & 24.86 / 0.7778 / 0.252 & 25.55 / 0.7805 / 0.285 \\
ZeDuSR-\(\ell\)                  & 24.98 / 0.7702 / 0.262 & \underline{26.16} / 0.7920 / 0.279 \\ \hline
KeDuSR-\(\ell\)               & \textbf{26.55} / \textbf{0.8325} / \textbf{0.186} & \textbf{27.24} / \textbf{0.8178} / \textbf{0.215} \\ \hline
\end{tabular}
}
\caption{Generalization evaluation with the model trained on DuSR-Real.}
  \label{tab:Generalization Comparison}
\end{table}

We also evaluate the generalization ability of different models trained on DuSR-Real by testing on the two other datasets. As shown in Table \ref{tab:Generalization Comparison}, our method has the best generalization ability, even outperforming the zero-shot learning method ZeDuSR. The main reason is that our kernel-free matching strategy is independent of cameras.   

\section{Conclusion}
\label{sec:Conclusion}
In this work, we proposed a KeDuSR network to deal with the real dual-lens SR task. We designed a global and local combined warping strategy to make the Ref well-aligned with the center region of LR input. Then, we formulate the LR center and the aligned reference as key-value pairs and propose a kernel-free matching strategy, whose matching index is used for corner warping. Afterward, we fuse the features of enhanced LR input and the features of corner and center well-aligned reference to generate the SR result. Experiments demonstrate the superiority of the proposed method. We also construct a DuSR-Real dataset with well-aligned pairs to facilitate research in this area.    

\clearpage
\clearpage

\section{Acknowledgments}
This work was supported in part by the National Natural Science Foundation of China under Grant 62072331, Grant 62231018, and Grant 62171317.

\bibliography{aaai24}

\clearpage
\clearpage

\section{Supplementary File}

\subsection{Real Dual-Lens Datasets}
The original RealMCVSR \cite{lee2022reference} dataset contains 161 sets of triples with ultra-wide, wide-angle, and telephoto videos, which can be used for $2\times$ and $4\times$ super-resolution with synthesized downsampling. Different from the original settings, we extract frames from the ultra-wide (LR) and wide-angle (HR and Ref) videos for 2$\times$ real-world image SR. To avoid repetitive content, for each video, we extract two or three frames. Then, we apply our alignment method (as mentioned in Fig. 1 of the main paper) on the extracted frames to construct LR-HR-Ref triples. After removing the LR-HR pairs with large misalignments, we construct the RealMCVSR-Real dataset, which contains 330 frames from 210 scenes. However, the scene number is still fewer than previous image SR datasets. Note that, the original RealMCVSR was captured with motion blur and the LR input is heavily degraded compared with the HR. In other words, the resolution gap between LR and Ref in this dataset is larger. Thanks to our kernel-free matching, our method has the best performance on this dataset and outperforms other methods by a large margin.

The original CameraFusion \cite{wang2021dual} dataset contains 146 pairs of wide-angle and telephoto images (4k resolution). However, the two kinds of images are captured by lens switching, leading to different contents between the two cameras in dynamic scenes. Therefore, we first remove the pairs that the contents cannot be matched in the overlapped FoV area. Then, we utilize our alignment method to construct LR-HR-Ref triples. After removing the LR-HR pairs with large misalignments, only 98 triples are kept. Note that, compared with the other two datasets, there are still small misalignments in this dataset due to its lens switching based capturing method.    

\begin{figure}[b]
  \centering

  \includegraphics[width=1\linewidth]{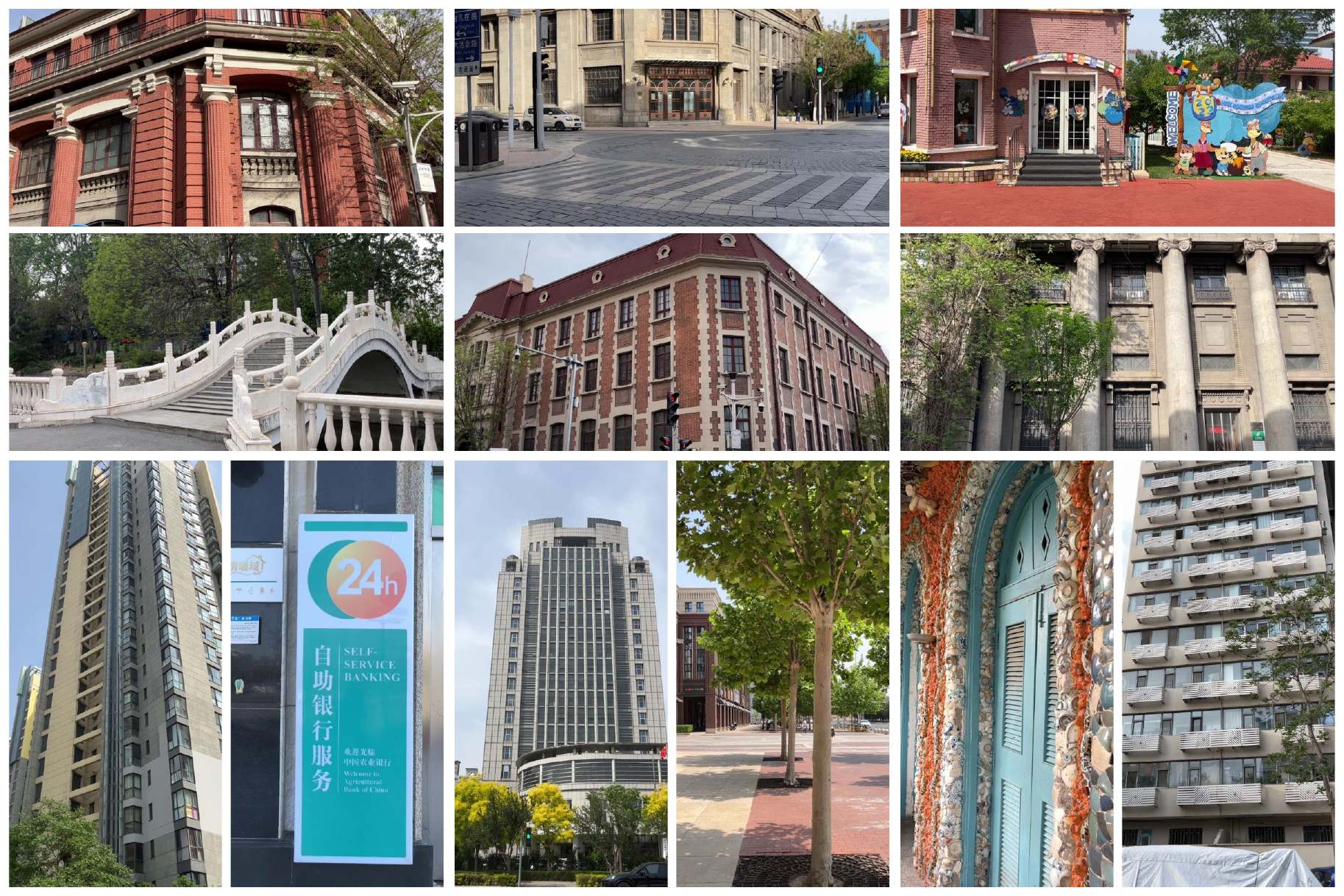}
   \caption{Exemplars of the aligned HR images in our DuSR-Real dataset.}   

   \label{examples}
\end{figure}

\begin{figure}
  \centering
  \includegraphics[width=1\linewidth]{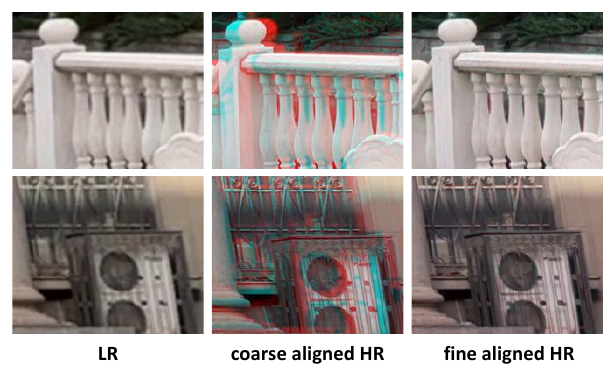}
  \vspace{-0.7cm}
   \caption{Illustration of LR-HR pairs in our dataset. From left to right: the LR input, the coarse aligned (SIFT based alignment) HR, and the fine aligned (deep flow based alignment) HR. The LR and fine aligned HR construct the LR-HR pairs in our dataset. To visualize the misalignments, we replace the red channel of the two kinds of HR by the red channel of the LR input. It can be observed that the coarse aligned HR has large displacements with the LR while the fine aligned HR is accurately aligned with the LR. }   
   \label{align}
\end{figure}

Considering the number of image triples in the above two datasets is still limited, we further construct our DuSR-Real dataset, which contains 420 training triples and 55 testing triples. Fig. \ref{examples} presents some examples of the HR images in our dataset, which cover multiple kinds of scenes. In addition, the LR-HR pairs in our constructed dataset are accurately aligned. As stated in the main paper, we utilize a coarse to fine alignment strategy. Fig. \ref{align} presents two examples of the input LR and its corresponding aligned HR. The first column is the input LR, namely the wide-angle image. The middle column is the coarse aligned HR. We replace its red channel by the red channel of the LR to visualize the misalignments between the two images. The right column is the HR after fine alignment. We also replace its red channel by that of the LR. It can be observed that after fine alignment, the LR and HR are well aligned, which constructs a suitable pair for supervised learning. 
Table \ref{Dataset} summarizes the three datasets. The image number and scene number in our dataset are the largest.

\begin{table}[h]
  \centering

\renewcommand\arraystretch{1.2}
  \resizebox{\linewidth}{!}{
\begin{tabular}{lcccc}
\hline
Dataset           & \begin{tabular}[c]{@{}c@{}}Image Number\\ (Training + Testing)\end{tabular} & \begin{tabular}[c]{@{}c@{}}Scene Number\\ (Training + Testing)\end{tabular} & \begin{tabular}[c]{@{}c}Scale \\Factor \end{tabular} & HR Resolution \\ \hline
DuSR-Real         & 420 + 55                                                                    & 305 + 55                                                                    & 2             & 1792 \(\times\) 896     \\
RealMCVSR-Real    & 330 + 50                                                                    & 210 + 27                                                                    & 2             & 1792 \(\times\) 896      \\
CameraFusion-Real & 83 + 15                                                                     & 65 + 15                                                                     & 2             & 3584 \(\times\) 2560     \\ \hline
\end{tabular}
  }
\caption{Comparison of the three datasets for real dual-lens SR.}
  \label{Dataset}
\end{table}

\subsection{Differences in Alignment for Dataset Construction and Center Warping}
For dataset construction (center warping), we utilize DeepFlow (flow-guided DCN). DeepFlow is more accurate but is non-differentiable and much slower than DCN. Therefore, for dataset construction, we utilize the more accurate Deepflow. By manually removing the cases that cannot be well aligned by DeepFlow, we generate a well-aligned dataset. For center warping, we utilize the faster and differentiable DCN.

\begin{table*}[!t]
  \centering

\renewcommand\arraystretch{1.2}
\resizebox{\linewidth}{!}{
\begin{tabular}{cccccccccccc}
\hline
           & RCAN & SwinIR & ESRGAN & BSRGAN & TTSR & MASA & DATSR & DCSR & SelfDZSR & ZeDuSR & KeDuSR \\ \hline
Latency(s) & 0.69 & 2.85   & 0.08   & 0.45    & 7.51 & 1.52 & 9.35  & 0.84 & 0.17     & 180    & 0.51   \\
Params (M) & 5.63 & 11.75  & 16.7   & 16.61  & 6.25 & 4.02 & 10.51 & 3.19 & 0.52    & 1.49  & 5.63  \\ \hline
\end{tabular}
  }
\caption{Latency and Parameter.}
  \label{tab:Latency}

\end{table*}

\subsection{Complexity Comparison.}
We present the latency and number of parameters in Table \ref{tab:Latency}. Our method ranks second among all the RefSR and the dual-lens SR methods in terms of latency. Latency indicates the time required to generate one HR result (1792 × 896) using one NVIDIA 3090 GPU.

\subsection{Visual Comparisons with State-of-the-arts}
\begin{figure*}
  \centering

  \includegraphics[width=1\linewidth]{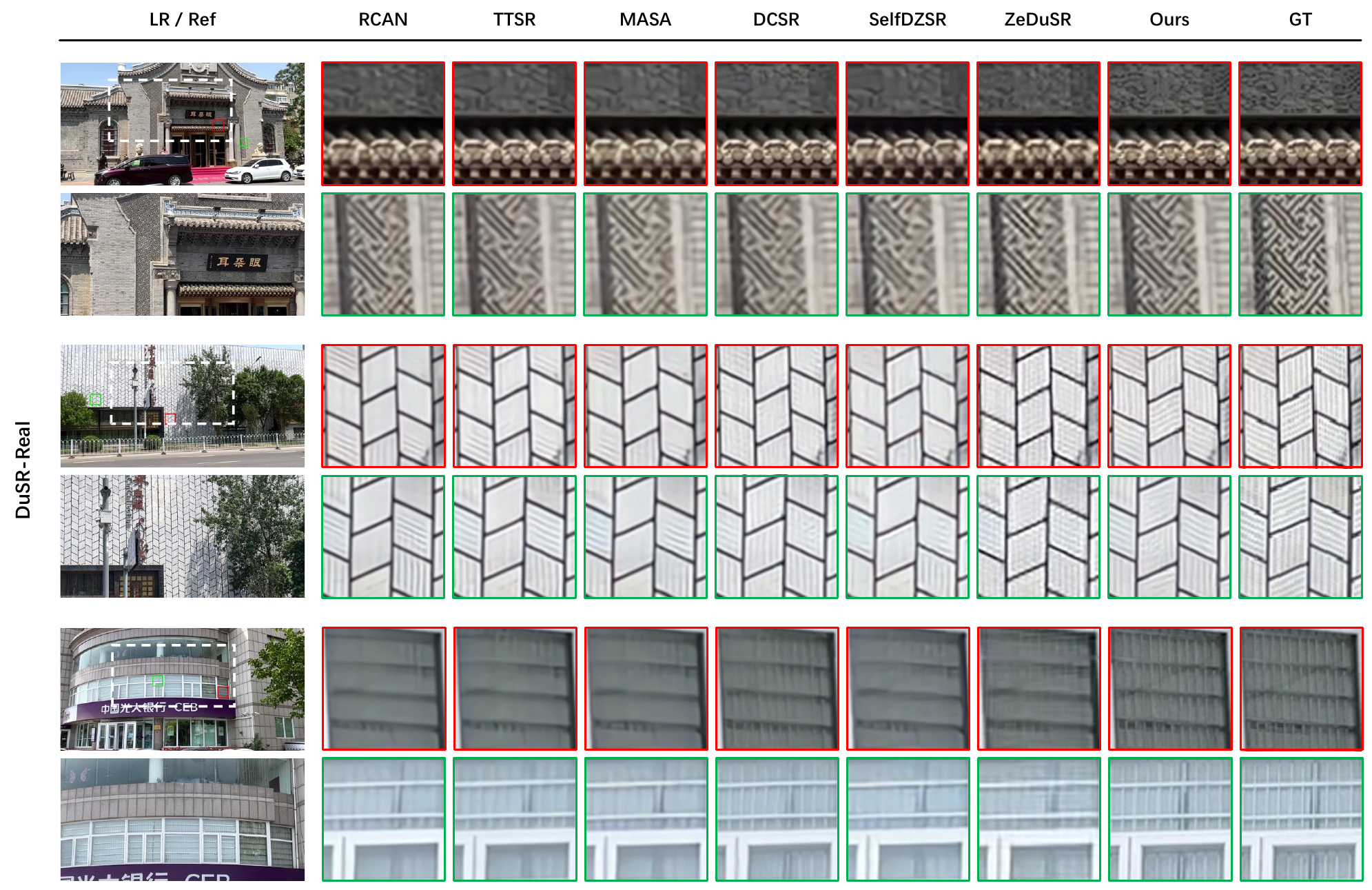}
   \caption{Visual comparisons on DuSR-Real dataset. The white dotted box indicates the overlapped FoV area between LR and Ref. The presented results are obtained with only reconstruction loss.}
   \label{visual1}
\end{figure*}

\textbf{Visual comparisons on DuSR-Real dataset.}
We further present more visual comparison results on DuSR-Real dataset. 
As introduced in the main paper, we compare with RCAN~\cite{zhang2018image}, TTSR~\cite{yang2020learning}, MASA~\cite{lu2021masa}, DCSR~\cite{wang2021dual}, SelfDZSR~\cite{zhang2022self}, and ZeDuSR \cite{xu2023zero}. Our approach demonstrates superior performance in both center and corner regions, as shown in Figs. \ref{visual1}.

\begin{figure*}
  \centering

  \includegraphics[width=1\linewidth]{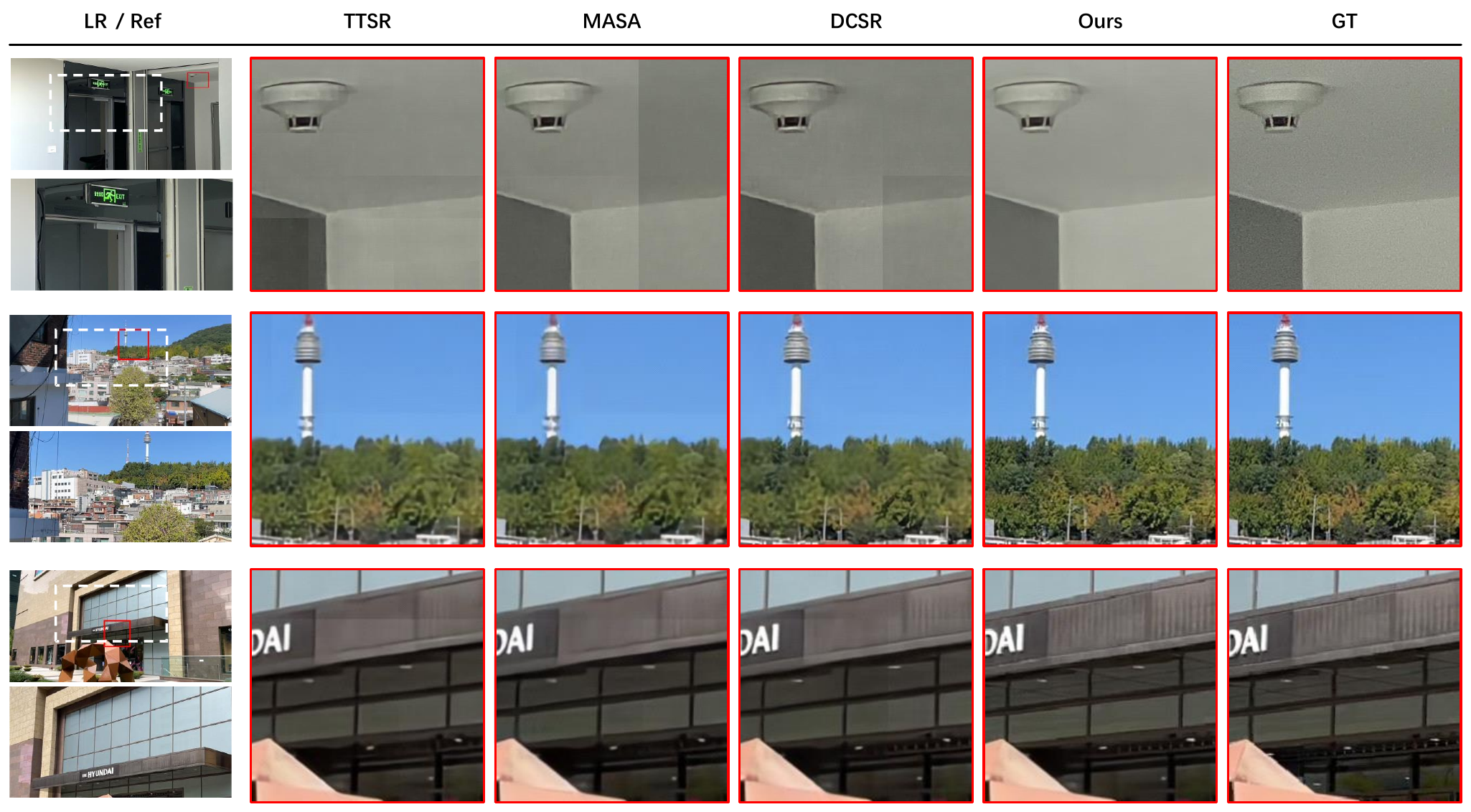}
   \caption{Visual comparisons about the blocking artifacts. The white dotted box indicates the overlapped FoV area between LR and Ref. The presented results are obtained with only reconstruction loss.}
   \label{blocking}
\end{figure*}

\textbf{Blocking artifacts.}
Both the RefSR and dual-len SR methods consume a substantial amount of GPU memory and we need to partition the LR input into blocks for processing, which may lead to blocking artifacts. In our implementation, we split the LR input into $128\times128$ patches with an overlap of 8 pixels. Except SelfDZSR and ZeDuSR, which consume small memory costs, all the other RefSR methods are split into patches in the same way. As shown in Fig. \ref{blocking}, TTSR, MASA, and DCSR introduce visible blocking artifacts. In contrast, our results have no blocking artifacts. 
The reason is that our approach utilizes kernel-free matching, which makes the brightness to be consistent across the whole image.

\begin{figure*}
  \centering

  \includegraphics[width=1\linewidth]{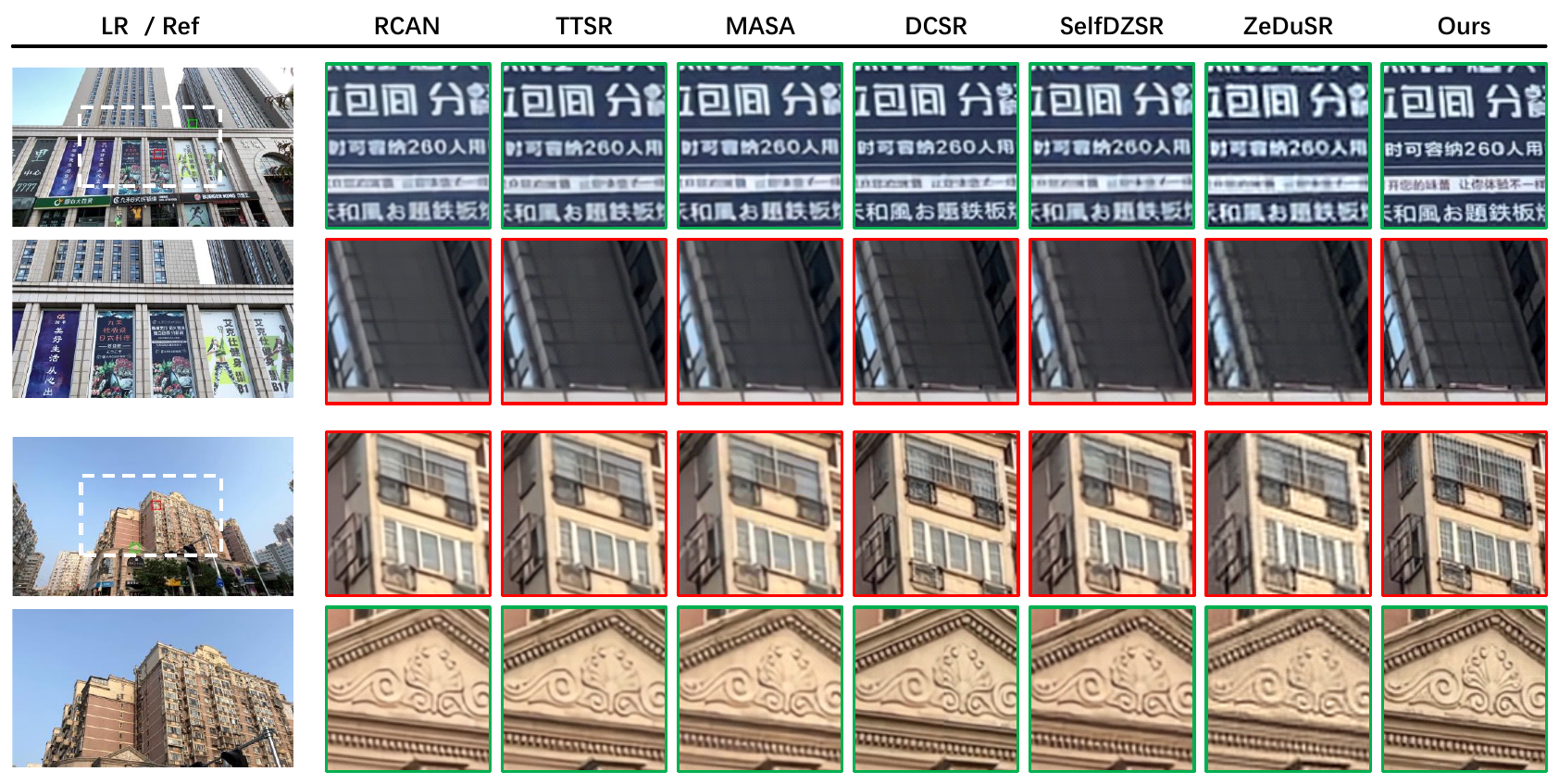}
   \caption{Visual comparisons on full-resolution inputs. The white dotted box indicates the overlapped FoV area between LR and Ref. The presented results are obtained with only reconstruction loss. Note that there is no ground truth in this case.}
   \label{full-resolution}
\end{figure*}

\textbf{Visual comparisons with full-resolution inputs.}
In order to have ground truth, the DuSR-Real dataset we have utilized for evaluation is constructed by cropped images. 
To comprehensively evaluate the effectiveness of our approach on real-world data, we utilize the real captured original full-resolution wide-angle and telephoto images as inputs for comparison, as shown in Fig. \ref{full-resolution}. Note that, in this case, there is no ground truth.  It can be observed that our method recovers the most vivid details for the center region. 


\subsection{Ablation Study Results}
\begin{figure*}
  \centering
  \includegraphics[width=1\linewidth]{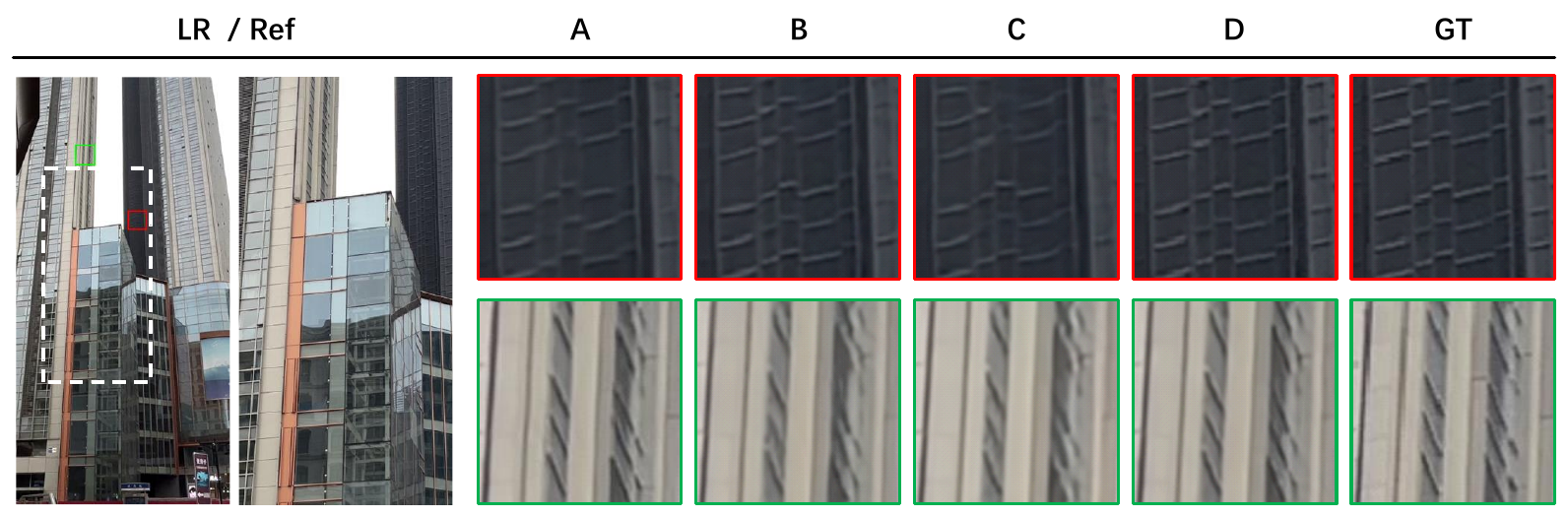}
   \caption{Visual comparisons of ablation study. The white dotted box indicates the overlapped FoV area between LR and Ref. The presented results are obtained with only reconstruction loss.}
   \label{Ablation Study}
\end{figure*}

\textbf{Visual comparisons.} As demonstrated in the main paper, there are four variants in our main paper. Since variant A utilizes traditional matching, its center and corner SR results are both worse than our full-model (as shown in Fig. \ref{Ablation Study}), namely variant D. For variant B, since there is no corner warping, its corner SR results are smooth. For variant C, since its center region is not well aligned, the center SR result is worse than our full model (D). Variant D achieves the best SR result in both center and corner regions.

\textbf{Ablation on global warping.}
As stated in the main paper, the first step for center warping is utilizing SIFT matching to calculate the homography transformation matrix. In this way, we align \(I^\text{Ref}\) with \(I^\text{IR}\)  globally (the global aligned Ref is \(\hat{I}^\text{Ref}\)), and identify the overlapped region \(I^\text{IRC}\). We would like to demonstrate that our method is robust to this step. Since the two cameras have a fixed relative position, we can also utilize a fixed homography transformation for the global warping.  
As indicated in Table \ref{tab:SIFT alignment}, `fixed' employs a fixed homography matrix to align each image globally, while `adaptive' indicates that each image is aligned using SIFT matching. The results demonstrate that even with `fixed' global transformation, our method still generates promising results. 

\begin{table}[h]
  \centering
\begin{tabular}{cc}
\hline
Global Warping  & DuSR-Real     \\ \hline
fixed                  & 27.51 / 0.8830 \\
adaptive               & 27.66 / 0.8890 \\ \hline
\end{tabular}
\caption{Ablation on the global warping method. The results are evaluated on our DuSR-Real dataset.}
  \label{tab:SIFT alignment}
\end{table}

\end{document}